# The generative paradox: the Anti-Prompt paradigm for human capability

*Position Paper*

**Margarita Trifonov[1]* and Vadim E. Levit[2]**

[1] Department of Science Teaching, Weizmann Institute of Science, Rehovot, Israel

[2] Department of Mathematics, Ariel University, Ariel, Israel

* Corresponding author: Margarita Trifonov, margarita.trifonov@weizmann.ac.il. Vadim E. Levit: levitv@ariel.ac.il.

**Abstract**

Generative AI in education is usually judged by the output of human-AI teams. That focus overlooks a second outcome: what people can still understand, challenge, repair, and adapt when assistance changes or disappears. We call this tension the generative paradox and distinguish four manifestations: performance, assistance, ownership, and capability stock. We propose developmental function allocation, which complements operational function allocation and meaningful human control by treating retained human capability as a design objective alongside current joint performance. The Anti-Prompt paradigm reverses the default interaction order: when immediate generation would preempt a consequential human act, AI should first create the conditions for human construction. We distinguish the validity of a product, responsibility for its construction, and evidence of capability after support changes. Representative Anti-Prompts, a canonical interaction sequence, and stress tests spanning mathematical reasoning, experimental science, software diagnosis, and lesson design make these distinctions concrete. We conclude with falsifiable predictions about retained capability and a system-level hypothesis about the reproduction and distribution of expertise.

## 1. The generative paradox

Generative AI is often evaluated by the quality and efficiency of the products that people and systems create together. Such gains matter for productivity and access (Noy & Zhang, 2023; Yan et al., 2024). Yet immediate joint performance is only one outcome of assistance. A second is what people can understand, challenge, repair, and reproduce after assistance changes or disappears.

As educational AI shifts from supporting discrete tasks to generating complete explanations, proofs, code, and lesson materials, the distinction between a polished product and the capability retained by the learner becomes a design problem for artificial intelligence in education.

Learning research has long distinguished performance during instruction from durable learning (Soderstrom & Bjork, 2015). Khosravi et al. (2026) bring this distinction to AI learning-companion design. Our focus is the allocation decision within an interaction: which consequential act should remain open to the learner, when should assistance disclose it, and what later evidence shows that the target capability has developed?

These outcomes can diverge. Across four experiments involving 3,562 participants, GenAI collaboration improved immediate professional output, but those gains did not consistently persist in subsequent solo work and were accompanied by lower intrinsic motivation and greater boredom (S.

Wu et al., 2025). In two logical reasoning studies involving 698 participants, AI assistance improved accuracy, whereas participants substantially overestimated their own performance (Fernandes et al., 2026). Product quality, independent capability, motivation, and calibrated self-knowledge can therefore move in different directions.

Evidence from STEM learning makes the same dissociation particularly visible. Unrestricted GenAI access improved assisted mathematics performance but reduced short-term unaided performance in one field experiment, while a teacher-informed guarded tutor largely mitigated this loss (Bastani et al., 2025). By contrast, a carefully designed AI tutor produced substantial learning gains in undergraduate physics (Kestin et al., 2025). In a recent preprint, developers assigned AI assistance while learning an unfamiliar programming library showed weaker subsequent conceptual understanding, code reading, and debugging, on average; exploratory analyses associated more cognitively engaged interaction patterns with better learning (Shen & Tamkin, 2026). The relevant question is not whether AI assistance is beneficial or harmful in general, but which functions it performs, which human operations remain available for practice, and which outcomes are measured and when (Soderstrom & Bjork, 2015).

We call the problematic case generative substitution: AI production displaces the human construction, evaluation, or revision through which a target capability is intended to develop or be maintained. Human responsibility alone does not make delegation substitutive. Delegating transcription, translation, or routine calculation may reduce incidental burden without displacing the capability at stake. Delegating proof strategy, causal diagnosis, model selection, or the choice of a lesson's central representation may instead remove the very operation through which future capability develops. The same AI contribution can therefore be peripheral for an expert, an essential accommodation for one learner, and substitutive for another. The paradox is relational, not a fixed property of a tool.

***The better AI becomes at producing intellectual products, the more deliberate education and professional formation should be in preserving opportunities to exercise and develop the capabilities needed to govern those products.***

### 1.1 Four manifestations of the generative paradox

The generative paradox appears in at least four forms. A performance paradox arises when better joint output coexists with weaker or unchanged independent capability. Unchanged independent performance is not, by itself, evidence of deterioration. Assisted improvement without learning, foregone learning relative to an appropriate alternative, and loss of a previously available capability require different comparisons. An assistance paradox arises when additional help improves progress while removing opportunities for generation, retrieval, selection, justification, or monitoring through which later human capability develops. An ownership paradox arises when visible authorship and consequential control diverge: a person may produce or edit the final artifact while AI has framed the problem, selected the strategy, or supplied the justification; conversely, AI may execute an operation while the person retains the capacity to challenge or override it and recover from its failure. A capability-stock paradox arises when locally efficient delegations accumulate in ways that narrow the future distribution of expertise needed to supervise AI-supported practices, recover from failures, teach successors, or improve those practices.

These are not logical contradictions but observable dissociations among outcomes measured at different times and levels of analysis. Conventional indicators such as output accuracy, time saved, user satisfaction, or the proportion of AI-generated content can conceal both who controlled the consequential act and what capability remains after support changes. This concern is also reflected in Agentivism, which similarly cautions against treating final product quality as sufficient evidence of learning (Yan & Gašević, 2026). Box 1 translates each manifestation into an observable dissociation and an empirical test, while Fig. 1 summarizes the underlying distinction between immediate joint performance and retained human capability**.**

Box 1. Four manifestations of the generative paradox

| Manifestation | Observable dissociation | Empirical test |
|---|---|---|
| **Performance paradox** | The joint product improves while independent capability does not improve or declines. | Compare immediate output with delayed reconstruction and transfer, using baseline and appropriate comparison conditions. |
| **Assistance paradox** | More help lowers friction but can remove generation, selection, justification, or monitoring. | Compare assistance conditions and code which cognitive opportunities remain available. |
| **Ownership paradox** | Human authorship can coexist with AI control of framing and strategy; AI execution can coexist with retained human capability. | Code framing, initiation, selection, execution, justification, monitoring, and authority to override and recover. |
| **Capability-stock paradox** | Local productivity rises while apprenticeship, recovery competence, or expertise diversity narrows. | Track opportunity, recovery, and expertise-distribution indicators longitudinally. |

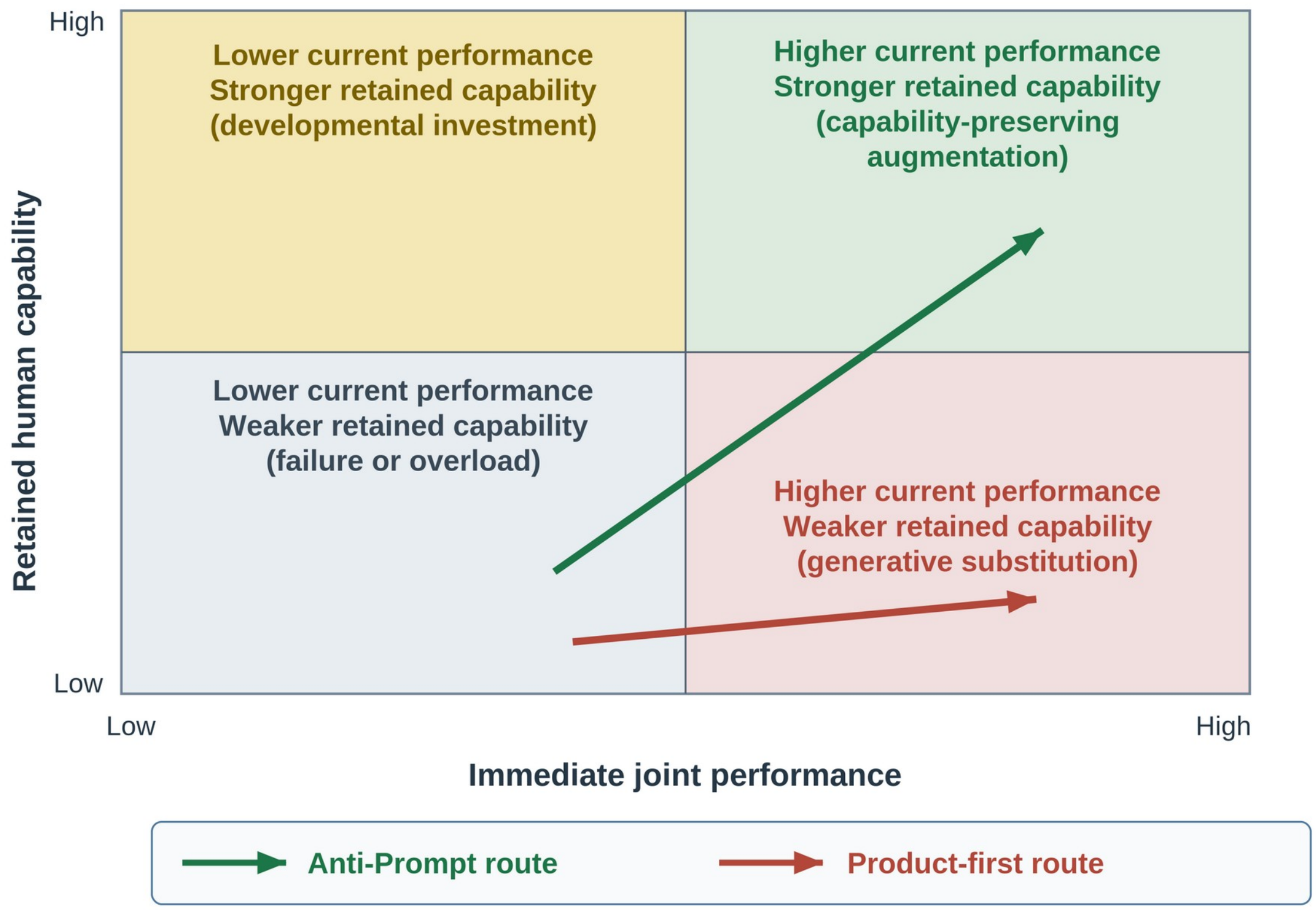

Fig. 1. The generative paradox. Improvement in immediate joint performance does not establish improvement in retained human capability. The arrows illustrate hypothesized Anti-Prompt and product-first routes, not exclusive outcomes or universal empirical laws. Comparisons require a stated baseline or alternative condition. Product-first assistance can support learning, and an Anti-Prompt intervention can fail. AI-assisted schematic: ChatGPT (OpenAI; GPT-6 Astra Pro for the current vector-code preparation) assisted with code and layout. Earlier model versions are not fully documented. The authors reviewed and edited the diagram; it depicts no empirical data.

## 2. Developmental function allocation

Allocating functions between people and machines is a longstanding problem in human factors research. Existing frameworks distinguish stages and levels of automation, treat allocation as context-dependent, and analyze the coordination and monitoring demands that arise when control is divided between human and automated agents (Dearden et al., 2000; Feigh & Pritchett, 2014; Li et al., 2013; Parasuraman et al., 2000; Pritchett et al., 2014a, 2014b). Bainbridge's classic "ironies of automation" identified a durable problem: people may remain responsible for exceptional failures after automation has reduced the opportunities through which they develop and maintain the expertise needed to detect and recover from those failures (Bainbridge, 1983).

Three strands of work converge on this problem. Human-factors research on function allocation addresses how tasks and control are distributed to support current performance, safety, workload, and situation awareness (Bainbridge, 1983; Dearden et al., 2000; Feigh & Pritchett, 2014; Li et al., 2013; Parasuraman et al., 2000; Pritchett et al., 2014a, 2014b). Research on learning, agency, and deskilling increasingly foregrounds how assistance can shape the opportunities through which human capability is developed, exercised, or lost (Chang et al., 2023; Ferdman, 2026; Ganuthula & Singh, 2026; Pelechano et al., 2026; Tsim & Gutoreva, 2026). Research on meaningful human control asks whether people retain the knowledge, authority, and capacity required for responsible oversight (Cavalcante Siebert et al., 2023; Davidovic, 2023). Developmental function allocation connects these strands by treating the capability produced or retained through today's allocation as an outcome of the allocation itself. It asks not only how work should be distributed now, but how that distribution changes what people will later be able to understand, challenge, recover, adapt, and teach.

Retained human capability is not a single score. It is a provisional profile across understanding, justification, monitoring and error detection, recovery skills, and adaptation or transfer. These dimensions need not move together. A person may understand an AI-supported result yet fail to reconstruct it. Capability should be assessed alongside the authority and practical opportunity to exercise it: detecting an error does not entail permission to override the system, and formal authority does not establish competence. Recent work on epistemic agency, epistemic co-agency, and the principle "scaffold, do not substitute" converges on the need to preserve active human reasoning (Favero et al., 2026; Samuel, 2026; J. Y. Wu et al., 2025).

Khosravi et al. (2026) organize AI learning companions around pedagogical, adaptive, and responsible-design foundations and explicitly call for evaluation of what learners can understand, transfer, and do independently after support is withdrawn. They also address which cognitive work should remain with the learner. Anti-Prompt can operate within such a companion: it connects a prospectively specified act, the timing and content of assistance, and later evidence of capability. The contribution is this integrated design and evaluation focus, not the invention of withholding, learner agency, or post-support testing. A learning companion can sustain the orientation across sessions; a single well-written prompt cannot guarantee that the interaction follows it.

The Agentivism preprint explains learning with AI through selective delegation, evaluation and reconstruction of AI contributions, and transfer under reduced support (Yan & Gašević, 2026). It also argues for preserving learner responsibility for framing and justification. Anti-Prompt is compatible with this account. We develop a specific design commitment: identify a consequential act prospectively, preserve a tractable opportunity to construct or justify it before disclosure, and test the resulting capability after support changes. The proposed comparison isolates the timing and content of assistance rather than treating learner agency or post-support assessment as newly introduced principles.

Operationalizing this framework requires consequential acts to be identified prospectively from the curricular or professional objective, expert task analysis, known novice failure modes, and the transfer or recovery demands of the practice. Different studies may legitimately protect different acts within the same task. The target and its prerequisites must be specified before learner responses or outcomes are observed. Responsibility is then represented as a time-varying profile across framing, initiation, selection, execution, justification, monitoring, and authority to override and recover. A protected opportunity is not evidence of successful participation: the record should distinguish an invitation to reason, an attempt, and a justified contribution, while allowing alternative valid solution paths. The framework asks three questions: Which act is being developed? Who or what supplies its consequential reasoning? What can the person do when equivalent assistance is unavailable? Capability dimensions should be tested as a possible common factor or task-specific profile; authority and opportunity should be recorded separately.

## 3. The Anti-Prompt paradigm

The Anti-Prompt paradigm translates developmental function allocation into an interactional design orientation. The anti-prompt is itself a prompt: it asks AI to preserve the conditions for human construction before producing the target. We use the term "paradigm" deliberately. The claim is not that questioning, scaffolding, or backward reasoning are new. It is that the default order of human-AI interaction should be reversed whenever product generation would preempt the practice through which the target capability develops. The design question shifts from "What should AI produce?" to "What human capability must remain?" Earlier AIED paradigm accounts distinguish AI-directed, AI-supported, and AI-empowered learning (Ouyang & Jiao, 2021). Anti-Prompt cuts across these roles by asking when AI should withhold a consequential step so that the learner can construct or justify it. Box 2 gives two representative requests.

Box 2. Two representative Anti-Prompts

**Mathematical proof.** *Help me prove that $\sqrt{2}$ is irrational, but do not give me the proof or the decisive contradiction. Begin by asking me how to represent the assumption that $\sqrt{2}$ is rational. Introduce the lowest-terms and parity ideas only if my responses show that I need them. After I finish, test what I can still do independently by asking why the same argument does not prove that $\sqrt{4}$ is irrational and how it changes for $\sqrt{3}$.*

**Experimental science.** *I want to test whether a pendulum's period depends on the attached mass. Do not tell me the answer or design the experiment. Ask me to state a prediction, choose the variable to change and the conditions to hold constant, and propose a measurement plan. Then raise one confound at a time. After I propose a design, ask which assumptions belong to the ideal model, which features of the real apparatus could make mass appear to matter, what evidence would count against my prediction, and how I would estimate uncertainty. Finally, ask me to adapt the design to test a different variable without further hints.*

The proof prompt is deliberately novice-facing and can provide less explicit guidance for advanced learners. It targets the justification of selected inferences rather than the unaided discovery of an entire proof strategy. Both requests are designed to preserve an opportunity for construction and to assess retained capability; the instructions alone do not establish that either occurred.

A backward-hurdle variant begins immediately before the finish line. It asks which penultimate claim would be sufficient for the conclusion, works backward through candidate subgoals, and then requires forward reconstruction without equivalent AI assistance. For example, consider an induction proof of the identity

$$1+2+\cdots+n=\frac{n(n+1)}{2}$$

The system can ask which algebraic form would yield the target formula for $k+1$, what justifies replacing the first k terms, and where the induction hypothesis is used. The reconstruction must include the base case and justify each step; a desired subgoal is not an established premise. Research on mathematical induction documents difficulties in coordinating these logical obligations (Arnesen & Skartsæterhagen, 2025), and proof-tutoring research cautions against treating backward subgoals as established facts (Matsuda & VanLehn, 2005). The Anti-Prompt contribution is not backward reasoning itself, but the regulation of disclosure followed by forward reconstruction.

The design logic identifies the consequential act, constrains premature AI production of that act, elicits an initial human model or strategy, provides contingent challenge and support, and returns control to allow retained human capability to be tested. Anti-Prompt is not anti-answer, anti-explanation, or anti-AI. Direct explanation may be appropriate when prerequisites are missing, continued struggle is unproductive, or safety, accessibility, or time constraints take precedence. Fig. 2 presents a canonical interaction sequence rather than a fixed protocol and allows branching toward prerequisite instruction, direct explanation, accessibility support, or durable hybrid work.

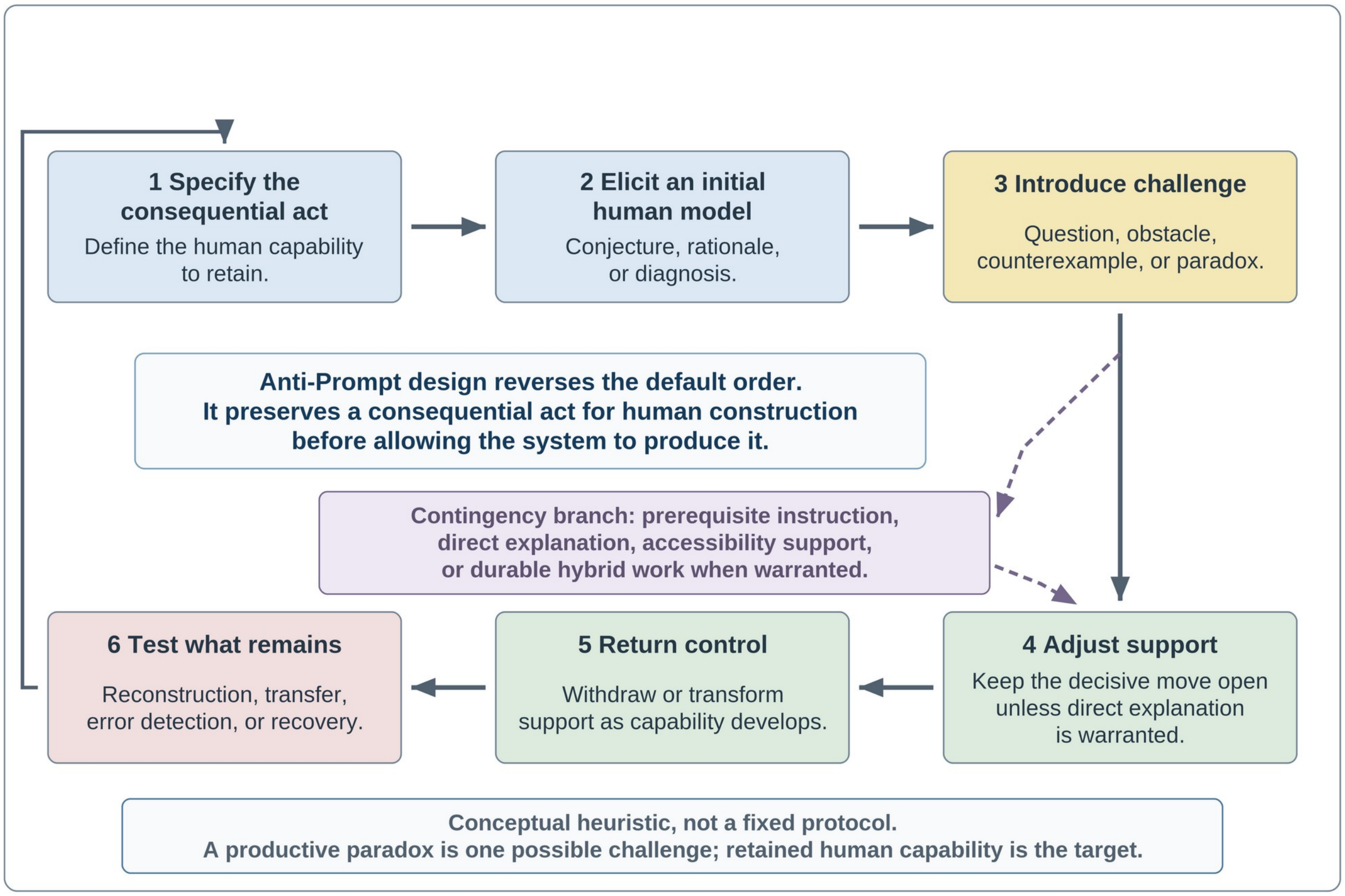

Fig. 2. A canonical Anti-Prompt sequence. The sequence specifies a consequential act, elicits a human model, introduces challenge, adjusts support, returns control, and tests what remains. A contingency branch allows prerequisite instruction, direct explanation, accessibility support, or durable hybrid work when warranted. The figure is a conceptual heuristic, not a fixed protocol. AI-assisted schematic: ChatGPT (OpenAI; GPT-6 Astra Pro for the current vector-code preparation) assisted with code and layout. Earlier model versions are not fully documented. The authors reviewed and edited the diagram; it depicts no empirical data.

Its psychological rationale concerns cognitive opportunities. Premature production can remove generation, retrieval, comparison among alternatives, commitment to a provisional model, and error-driven revision. Such delegation can function as cognitive offloading, shifting parts of the task from internal processing to an external aid (Risko & Gilbert, 2016). Fluent AI output can also be mistaken for evidence of one's own understanding, weakening confidence-accuracy calibration and reducing the perceived need for further checking (Tankelevitch et al., 2024). Different forms of assistance can displace different operations: complete answers can remove generation and retrieval; highly directive hints can remove selection; authoritative rationales can remove justification; and automatic evaluation can remove monitoring.

Learning research provides both mechanisms and limits. Self-explanation and constructive engagement can elicit inferences that passive inspection does not (Chi et al., 1989; Chi & Wylie, 2014). Productive failure can support later consolidation when tasks are tractable and feedback is well timed (Kapur, 2014; Loibl & Rummel, 2014). Scaffolding and the assistance dilemma, however, show that withholding is not intrinsically beneficial (Koedinger & Aleven, 2007; van de Pol et al., 2010; Wood et al., 1976). Novices may need explicit modeling before independent work, while support that benefits novices can become redundant for experts (Kalyuga et al., 2003; Renkl & Atkinson, 2003). Anti-Prompt therefore seeks the least substitutive assistance consistent with progress, access, and responsible performance, not the least assistance possible.

The same concern extends beyond formal learning. Repeated exposure to biased AI can alter later independent judgments (Glickman & Sharot, 2025), AI-supported science can produce illusions of understanding (Messeri & Crockett, 2024), and human-AI combinations do not consistently outperform their strongest member (Vaccaro et al., 2024). Cognitive-forcing interventions can reduce overreliance, sometimes at the cost of convenience (Buçinca et al., 2021), while generative tools can increase individual creativity yet narrow collective diversity (Doshi & Hauser, 2024). These findings reinforce the central dissociation: gains in one visible outcome can coexist with losses in another.

Questioning alone is therefore insufficient. A model can ask leading questions that narrow the option space so tightly that the learner merely executes an AI-framed solution. Conversely, a delayed worked example, proof checker, simulator, or test generator may preserve capability when introduced after a meaningful human attempt. Cognitive apprenticeship, educational reverse prompting, and Socratic LLMs provide important precedents (Chang et al., 2023; Collins et al., 1989; Ding et al., 2024; Liu et al., 2024). The distinctive Anti-Prompt claim is that such mechanisms should be organized around a protected consequential act and evaluated by what the person can still reconstruct, challenge, transfer, or recover after equivalent support is no longer available.

Consider two learners who finish the same correct proof. One proposes and justifies a divisibility inference; the other receives the inference inside a leading question and writes it down. The products alone cannot distinguish these interactions. Nor does the record establish that the first learner retained more: who first states a step is not a complete measure of who understands it. The comparison must record what each learner was invited to do, what assistance supplied, and what each can later justify or adapt. Changing an answer into a question does not necessarily return the reasoning to the learner.

## 4. When paradox makes revision necessary

Paradox can serve as one form of productive challenge within the Anti-Prompt paradigm. The term operates at two levels. The generative paradox describes a dissociation between immediate joint performance and retained human capability. A pedagogical paradox is a counterintuitive result or boundary case that exposes the limits of a learner's current model.

The Koch snowflake provides a transparent illustration. As the construction is iterated, its perimeter grows without bound while its enclosed area converges to a finite value. A learner who expects perimeter and area to move together must identify the failed assumption and construct a more precise account of why perimeter and area have different convergence behavior. The pedagogical value lies not in surprise itself, but in the conceptual revision required to explain it. Research on proof comprehension and GenAI-assisted proving similarly emphasizes structure, dependency, and critical validation rather than local correctness alone (Mejía-Ramos et al., 2012; Selden & Selden, 2003; Stylianides et al., 2024; Yoon et al., 2024; Yoon et al., 2026; Zhuang, 2026).

Generative AI can preempt this opportunity by presenting both the anomaly and its resolution before the learner has committed to an interpretable expectation. An Anti-Prompt sequence instead elicits a provisional model, introduces the smallest anomaly or boundary case that the model cannot explain, and asks the learner to diagnose the failure before supplying the decisive distinction. Contradictory information alone is insufficient: learners may discount, reinterpret, or compartmentalize it unless a coherent alternative becomes available (Potvin, 2023).

The resulting checkpoint has four functions: commitment to an explicit expectation, exposure to a minimally sufficient anomaly, diagnosis of the failed assumption, and reconstruction tested against a neighboring case. The protected act is not the experience of surprise but the work of conceptual revision. An anomaly is productive only if it exposes a misconception or model boundary that matters to the practice being developed. Generic surprise is not enough.

## 5. Stress tests across STEM practice

The proof case is deliberately compact. In a proof of the irrationality of $\sqrt{2}$, the protected act may be identifying the contradiction or justifying the parity step rather than reproducing the completed proof. The Box 2 prompt is novice-facing: it signposts part of the architecture while preserving execution and justification; an advanced version can withhold the lowest-terms and parity cues (Mejía-Ramos et al., 2012; Selden & Selden, 2003; Stylianides et al., 2024; Yoon et al., 2024; Yoon et al., 2026; Zhuang, 2026). In experimental science, the pendulum prompt protects prediction, variable control, and the distinction between an ideal model and a real apparatus.

Software incident response provides a professional test. The visible product may be a successful patch, while the consequential act is causal diagnosis: identifying the failure mechanism, tracing dependencies, and selecting a safe recovery strategy. An AI-generated fix can restore service without preserving that capability. Stronger evidence of retained human capability is the ability to explain the causal chain, reject a superficially effective but unsafe intervention, and recover from a related failure under altered conditions.

Lesson design provides a teacher-side test. GenAI can generate pedagogical architecture, not merely format materials. Studies show variation in whether teachers adopt, contextualize, reject, or revise model suggestions during AI-supported planning (Choi et al., 2026; Krushinskaia et al., 2026; Velander, 2026). Research on pedagogical reasoning, design capacity, and technology orchestration treats teaching as the adaptation and transformation of resources rather than their passive use (Beyer & Davis, 2012; Brown, 2009; Drijvers et al., 2010; Shulman, 1987). An Anti-Prompt interaction protects consequential design decisions, such as the target idea, anticipated student thinking, representation sequence, and evidence of learning, before materials are generated. Here, too, polished output is insufficient evidence of ownership; stronger evidence lies in justified rejection, counterfactual adaptation, and enactment under changed conditions.

## 6. The capability-stock hypothesis

The capability-stock manifestation motivates a system-level hypothesis: repeated allocation decisions change not only present products but also the future distribution of expertise. We call this the capability-stock hypothesis**.** It is not yet a validated stock measure. It proposes that opportunities to practice consequential acts are inflows to future capability, while nonuse, attrition, and organizational exit are sources of depreciation. Local efficiency can therefore diverge from long-term resilience.

Possible indicators include the number and distribution of independently capable practitioners, recovery performance after automation failure, training throughput, the breadth of unaided alternatives generated, and succession or apprenticeship rates. An observational analysis of AI adoption in science found higher individual output alongside smaller teams, fewer junior researchers, and a narrower collective research focus (Hao et al., 2026). These associations do not establish deskilling, but they are consistent with a possible bottleneck in the reproduction and distribution of expertise. The capability-stock hypothesis should therefore be tested longitudinally rather than inferred from output alone.

This is also an equity problem. If consequential decisions remain largely the domain of established experts while novices are confined to inspecting finished AI products, automation may concentrate rather than democratize opportunities to develop expertise. A capability-preserving institution must consider access to formative practice, not merely access to AI. The objective is not to preserve manual performance for its own sake, but to maintain a sufficiently broad distribution of human capability for independent challenge, recovery, teaching, and innovation wherever those capacities remain consequential.

## 7. What would count as a decisive test?

A strong first comparison would contrast direct AI generation, a complete AI-generated solution followed by structured self-explanation, high-quality Socratic assistance, and Anti-Prompt allocation. Consistent with Khosravi et al. (2026), outcomes should distinguish immediate joint performance from delayed reconstruction, near and far transfer, detection of unfamiliar errors, confidence calibration, and recovery. The act-level question is which parts of construction, evaluation, or revision were supplied by the learner, AI, or other tools. Comparisons should distinguish those actual allocations rather than rely on labels such as "Socratic" or "Anti-Prompt."

The Austin-Law Proof-Synthesis (ALPS) preprint offers a methodological contrast, not evidence about human learning. Xie et al. (2026) check both whether submitted equations entail a target law and whether they admit a structure with at least two distinct elements. In the separate proof channel, supporting software may supply short missing derivations before the completed proof is checked (Appendix B.2 and Figure A1). This distinction matters educationally: verifying a submitted step is different from supplying one. We therefore propose recording tool-generated reasoning alongside learner and AI contributions, even when that reasoning is absent from the conversational transcript. This is an educational inference from the benchmark design, not an ALPS learning result.

A focused study could randomly assign learners with the specified prerequisites to attempt a target inference before disclosure or to receive it before explaining it. The task family, model, final explanatory material, correctness checker, and scheduled support budget would be matched; actual assistance and time used would be recorded. After a prespecified delay, independent assessors would score learners' responses to a prevalidated task variant completed without assistance that supplies the target reasoning. The primary analysis should follow the assigned conditions and include learners with incomplete products, rather than only successful submissions. Attrition should be reported explicitly. Product success, later capability, and instructional costs should be reported together. The test concerns what the allocation changes, not which condition produces more attractive dialogue.

No single omnibus comparison can identify every active ingredient. Component studies can separately vary construction protection, visibility of responsibility, and the withdrawal or transformation of support; a further study can compare forward and backward hurdle sequencing. Primary outcomes and scoring rules should be set in advance, with scorers unaware of condition where feasible. Analyses should distinguish total intervention effects from pathways through engagement and preserved opportunities for generation, selection, justification, and monitoring. A statistical mediation result alone would not establish a causal mechanism. Responsibility profiles should be coded independently of the outcome assessments.

The central hypothesis is conditional: for learners with the required prerequisites, preserving a tractable opportunity to construct or justify the target act will improve later performance on that act relative to matched early disclosure. The expected benefit should be smaller or absent when the act is already mastered or when indispensable prerequisites are missing (Kalyuga et al., 2003; Renkl &

Atkinson, 2003). These are predictions to test, not observed Anti-Prompt effects. Predictive analyses of responsibility profiles should remain separate from estimates of intervention effects. A study should prespecify the smallest educationally meaningful advantage and report uncertainty; a nonsignificant difference alone does not establish equivalence. An estimate precise enough to exclude the predicted meaningful benefit would count against that prediction. Implementation-level superiority requires comparison with strong Socratic assistance (Ding et al., 2024; Liu et al., 2024), independent-first use (Yoon et al., 2026), or other capability-preserving approaches (Chan, 2026; Chen, 2025; Ramos-Benitez et al., 2026; Shintani, 2026; Vendrell & Johnston, 2026). Equivalent outcomes between interventions that preserve the same acts establish no superiority of either implementation. The paradigm must survive comparison, not merely attract attention.

## 8. Assessment and governance

**A verified product is not a verified learner.** A proof certificate supports a formal claim under its checking assumptions; it does not by itself establish a person's understanding, the originality of the reasoning, or who supplied a decisive step. The ALPS example makes the limits of this inference concrete (Xie et al., 2026). Three questions therefore require different evidence: Is the product valid? Who or what supplied the consequential reasoning? What capability remains after support changes? Fresh task variants can reduce exact-answer rehearsal, but a new instance does not guarantee new reasoning. Educational variants need verified solutions, appropriate difficulty, and alignment with the protected act. Assessment should sample delayed reconstruction, unfamiliar errors, recovery from misleading suggestions, and adaptation to changed constraints. Removing equivalent reasoning assistance need not remove accessibility accommodations or ordinary tools that do not perform the target act. No single indicator is decisive: reconstruction can be rehearsed, confidence can be miscalibrated, and participants can shape interaction traces strategically.

Governance should likewise distinguish product-oriented from development-oriented AI use. Policies should specify where direct generation is appropriate, where capability-preserving allocation is warranted, which accessibility or safety considerations override default constraints, and who retains responsibility when AI guidance fails. Interfaces could display current allocations and allow users to reframe or override them. Research records should capture the assistance needed to interpret the specified target act, not an unrestricted record of learner activity. Privacy, transparency, and human agency remain central constraints (Kasneci et al., 2023; OECD, 2026; UNESCO, 2024).

The aim is not to preserve effort, manual performance, or existing occupational boundaries for their own sake. It is to preserve a sufficiently broad distribution of human capability for informed delegation, independent challenge, recovery, teaching, and innovation. This objective is compatible with hybrid intelligence and shared epistemic agency: consequential work can be distributed across people and tools without requiring every operation to remain human, provided that responsibility remains visible and recoverable (Borba et al., 2026; Damşa et al., 2010; Molenaar, 2022).

## 9. Conclusion

Generative AI creates a paradox that better output alone cannot resolve: the same assistance that improves present performance can displace the practice through which future human capability develops. Developmental function allocation makes retained human capability a design outcome alongside current joint performance. The Anti-Prompt paradigm translates that principle into a reversal of the default order: when immediate generation would substitute for a consequential human act, AI should first create the conditions for human construction.

The most capable AI need not always act first. In some contexts, its most valuable contribution is a question, an obstacle, a counterexample, or a paradox that leaves the consequential act open to human construction. The joint product is one outcome of human-AI interaction. Retained human capability after assistance changes is a second outcome. In AI-enabled learning environments, and in the professional practices that such environments prepare learners to enter, both should be designed for, measured, and valued.

## Data availability

No new datasets were generated or analyzed for this Position Paper.

## Declaration of generative AI and AI-assisted technologies in the manuscript preparation process

In preparing this work, the authors used ChatGPT (OpenAI) to assist with literature discovery, comparison and synthesis of sources, structural and language revision, and code-based preparation of explanatory figures. The authors independently verified the cited sources, reviewed and edited all AI-assisted material, and take full responsibility for the content of this Position Paper.